\documentclass[letterpaper, 10 pt, conference]{ieeeconf}

\usepackage[T1]{fontenc}

\IEEEoverridecommandlockouts
\usepackage{amsmath, amssymb}
\usepackage{mathtools}
\usepackage{bm}
\usepackage{graphicx}
\usepackage{dsfont}
\usepackage{float}
\usepackage{amsfonts}
\usepackage{ragged2e}
\usepackage{caption}
\usepackage{dsfont}
\usepackage[ruled,linesnumbered]{algorithm2e}

\SetCommentSty{mycommfont}

\usepackage{hyperref}
\usepackage{booktabs}
\usepackage{array}
\newcolumntype{$}{>{\global\let\currentrowstyle\relax}}
\newcolumntype{^}{>{\currentrowstyle}}
 
\usepackage{bm}
\newcolumntype{L}[1]{>{\raggedright\let\newline\\\arraybackslash\hspace{0pt}}p{#1}}
\newcolumntype{C}[1]{>{\centering\let\newline\\\arraybackslash\hspace{0pt}}p{#1}}
\newcolumntype{R}[1]{>{\raggedleft\let\newline\\\arraybackslash\hspace{0pt}}p{#1}}
\usepackage[x11names]{xcolor}
\usepackage{subcaption}

\newcommand{\ra}[1]{\renewcommand{\arraystretch}{#1}}
\usepackage{siunitx}

\title{Adaptive Splitting of Reusable Temporal Monitors for Rare Traffic Violations}

\author{Craig Innes and Subramanian Ramamoorthy$^{1, 2}$%
\thanks{Authors from the University of Edinburgh,
EH8 9AB, United Kingdom. Corresponding author: craig.innes@ed.ac.uk. Work supported by a grant from the UKRI Strategic Priorities Fund to
the UKRI Research Node on Trustworthy Autonomous Systems Governance
and Regulation (EP/V026607/1, 2020-2024). For the purpose of open access,
the author(s) has applied a Creative Commons Attribution (CC BY) license
to any Accepted Manuscript version arising}
}

\begin{document}

\maketitle

\begin{abstract}

Autonomous Vehicles (\textsc{av}s) are often tested in simulation to estimate the probability they will violate safety specifications. Two common issues arise when using existing techniques to produce this estimation: If violations occur rarely, simple Monte-Carlo sampling techniques can fail to produce efficient estimates; if simulation horizons are too long, importance sampling techniques (which learn proposal distributions from past simulations) can fail to converge. This paper addresses both issues by interleaving rare-event sampling techniques with online specification monitoring algorithms. We use adaptive multi-level splitting to decompose simulations into partial trajectories, then calculate the distance of those partial trajectories to failure by leveraging robustness metrics from Signal Temporal Logic (\textsc{stl}). By caching those partial robustness metric values, we can efficiently re-use computations across multiple sampling stages. Our experiments on an interstate lane-change scenario show our method is viable for testing simulated \textsc{av}-pipelines, efficiently estimating failure probabilities for \textsc{stl} specifications based on real traffic rules. We produce better estimates than Monte-Carlo and importance sampling in fewer simulations.
\end{abstract}

\section{Statistical Simulation for AV Testing}


Autonomous Vehicles (\textsc{av}s) typically undergo rigorous simulated testing before deployment \cite{rajabli2020software}. A standard set of steps for testing is as follows: First we define a scenario (e.g., a highway lane-change expressed in OpenScenario \cite{riedmaier2020survey}). Next, we define a safety specification (e.g., \emph{``avoid impeding traffic flow''}) in a formal language like Signal Temporal Logic (\textsc{stl}). Then, we run stochastic simulations to estimate the probability our \textsc{av}-system violates our specification \cite{corso2021survey}. This \emph{statistical simulation} approach is used because modern \textsc{av}s contain ``black box'' components like Neural-Network perception modules and non-linear solvers. Such components provide few analytical guarantees over their behaviour.

A core problem plaguing statistical simulation is estimating \emph{rare events}. Consider a stochastic simulation scenario where there exists a $10^{-4}$ probability that random noise in the sensors will cause our \textsc{av} to ``fail'' (i.e., to violate our safety specification). If we ran $100$ simulations, it is likely none would produce a failure. Even if sampling did produce a failure, estimation variance would be unacceptable \cite{juneja2006rare}.

Many works address rare-event problems for \textsc{av}s via \emph{Importance Sampling} \cite{bugallo2017adaptive}: Importance samplers draw simulations from a proposal distribution where the factors leading to a failure occur more frequently. The final estimate of failure probability is then re-weighted to reflect the original distribution. Since we do not know in advance all combinations of states which result in failure, such techniques must \emph{learn} a good proposal. This learning step has no convergence guarantees, and probability estimates from such adaptive techniques can have unbounded error \cite{arief2021certifiable}. Importance sampling also tends to fare better when failures are caused by instantaneous single-state errors, but in the \textsc{av} domain, failures often occur as a result of accumulated errors over dependent states \cite{cerou2019adaptive}. 

\begin{figure}
\centering
    \includegraphics[width=0.9\linewidth]{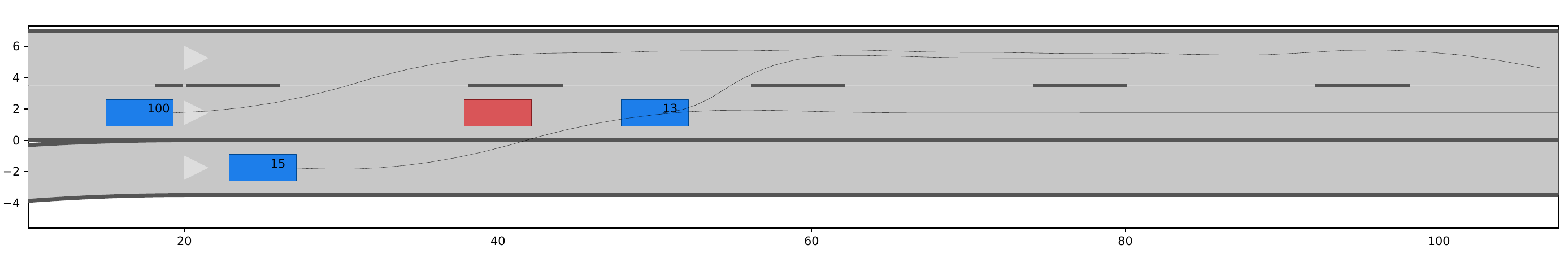}
    \centering
    \color{Firebrick4}{\textbf{\emph{``Preserve Traffic Flow''}}}
    \begin{equation*}
    \centering
    \color{darkgray}
    \scriptsize
    \square_{[0, \infty]}(\neg \text{slow\_leading\_vehicle}(x^{\textit{ego}}, x^{o_1 \dots o_3})\implies \text{preserves\_flow}(x^{\textit{ego}})) 
    \end{equation*}
\caption{Lane-change. Moving vehicles (blue) shown with trajectory. `Ego' vehicle must avoid static obstacles (red). We monitor the safety constraint shown in English and \textsc{stl}.}
\label{fig:scenario}
\end{figure}



This paper instead proposes an approach to \textsc{av} rare-event simulation based on merging \emph{Adaptive Multi-level Splitting} (\textsc{ams}) \cite{cerou2019adaptive} with \textsc{stl} monitoring. \textsc{ams} relies on estimating probabilities for a sequence of decreasing failure thresholds $\gamma_{1} > \gamma_{2} \dots > \gamma_{M}$, where the final $\gamma_M$ is equivalent to the rare failure event of interest. The key idea is that estimating any intermediate $\gamma_{i}$ (given $\gamma_{i-1}$) is easier than estimating $\gamma_M$ outright. To adapt \textsc{ams} from estimating isolated phenomena (e.g., particle transport \cite{louvin2017adaptive}) to estimating complex \textsc{av}-system failures, we face two issues:

The main issue is how to consistently produce simulations which fall below those intermediate failure thresholds $\gamma_{0 \dots M}$, and how to efficiently measure the distance to failure in the first place. Our approach measures failure using metrics for evaluating \textsc{stl} specification robustness. By leveraging online monitoring \cite{deshmukh2017robust}, we can cache the robustness values of partial trajectories, stop simulations at the point where they fall below the current threshold, and re-sample from this point onwards to produce trajectories which fall below subsequent failure thresholds.

A secondary issue is generating stochastic \textsc{av} perceptual errors. Approaches which assume noise follows a well-known (e.g., Gaussian) state-independent distribution \cite{carvalho2014stochastic} are insufficient to capture the perceptual variety of a typical \textsc{av}-system---a LiDAR detector may be great for close range traffic, but terrible for long range or occluded traffic \cite{pandharipande2023Sense}. We therefore use a \emph{Perception Error Model} (\textsc{pem}) \cite{piazzoni2023pem}---a surrogate trained on real sensor data which mimics perceptual errors encountered in regular operation (Sec \ref{sec:method:pems}).

The main contribution of this paper is a new method for assessing failure probability in \textsc{av}-scenarios (Sec \ref{sec:method}), combining \textsc{ams} (Sec \ref{sec:method:ams}), \textsc{pem}s (Sec \ref{sec:method:pems}), and online \textsc{stl} monitoring (Sec \ref{sec:method:online-stl}). Our experiments focus on the case study of a highway lane-change scenario, and show our method can be used to test a full \textsc{av}-pipeline---perception down to control (Sec \ref{sec:exp:setup}). Our approach outperforms Monte-Carlo sampling, as well as fixed and adaptive importance sampling, across various \textsc{stl} specifications (Sec \ref{sec:exp:results}). 

To limit the scope of experiments, this paper exclusively considers \emph{probabilistic noise in the perception system} as the primary source of simulation stochasticity (As is standard in other works \cite{corso2020scalable}). However, our proposed sampling method can easily be applied to simulators which consider other sources of stochasticity such as those arising from traffic behaviour or physical uncertainties. 

\section{Probability of Failure in Black Box Simulation}
\label{sec:problem-statement}

Consider the lane-change maneuver in Fig (\ref{fig:scenario}). Our car (the left-most \emph{ego vehicle}), must change to the left lane to avoid an obstacle, then re-merge. Formally, let's assume our scenario takes place over a total of $T$ time steps. We denote the $d$-dimensional state of our ego vehicle at time $t$ as $x^{ego}_{t} \in \mathds{R}^d$; other vehicles as $x^{o_i}_{t}$. For succinctness, we write $x_t = \langle x^{ego}_t, x^{o_0}_t, \dots, x^{o_M}_t\ \rangle$ for the combined state. The state $x_t$ contains the position, velocity and rotation of each vehicle. At each time step $t$, the ego vehicle's control system takes an action $a_t \in \mathds{R}^2$ (desired acceleration and turn-velocity) with the aim of minimizing costs associated with competing driving goals (e.g., maintaining a reference velocity and minimizing abrupt steering), and subject to constraints (e.g., limits on acceleration, avoiding collisions, staying within road boundaries). We can run a \emph{simulation} of this system to generate a trajectory $\tau=[(x_0, a_0) \dots (x_T, a_T)]$, where $\tau_{[i{:}j]} = [(x_i, a_i) \dots, (x_j, a_j)]$) denotes a partial slice. For a given scenario, we wish to test whether our above \textsc{av}-system will violate an \textsc{stl} safety specification $\varphi$. Due to probabilistic noise in the perception system, our simulator is inherently stochastic. Therefore our aim is to calculate the probability that, for a random run of our simulator, our \textsc{av}-system will violate $\varphi$: 

\begin{equation}
    P_{\textit{fail}} = \mathds{E} \left[ \mathds{1} \{ \tau \nvDash \varphi \} \right]
    \label{eqn:prob-failure-goal}
\end{equation}
where $\mathds{1} \{ \tau \nvDash \varphi \}$ is an indicator function which returns $1$ if $\tau$ violates $\varphi$ and $0$ otherwise. To explain how our method efficiently calculates (\ref{eqn:prob-failure-goal}), we first cover the pre-requisites for perception, tracking, and control for simulating our \textsc{av} (Sec \ref{sec:method:pems}-\ref{sec:method:mpc}). We then cover defining safety specifications $\varphi$ in \textsc{stl}, and how to quantify their satisfaction using a robustness metric (\ref{sec:method:stl}). We can then describe our main contribution---interleaving online monitoring and Adaptive Multi-level Splitting to estimate a failure probability for \textsc{av}-systems via statistical simulation (\ref{sec:method:ams}).

\begin{figure}
    \centering
    \includegraphics[width=0.9\linewidth]{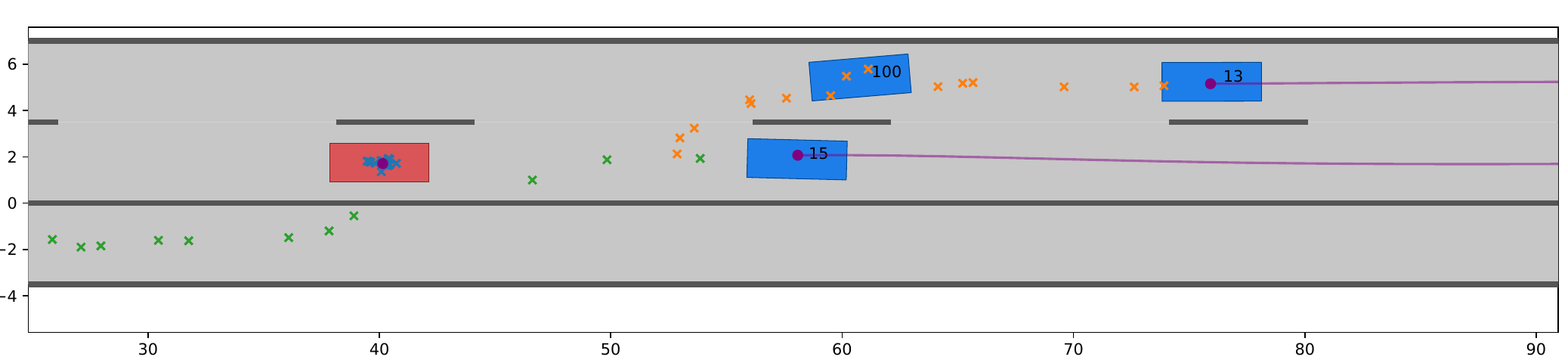}
    \caption{Lane change with \textsc{pem} observations, tracking, and prediction. Green/Orange crosses show \textsc{pem} obstacle observations. Purple dots/lines mark estimated/predicted positions.}
    \label{fig:scenario-with-tracking}
\end{figure}

\subsection{Simulated State Estimation with PEMs}
\label{sec:method:pems}

Most \textsc{av} problems assume our system does not have access to the true state $x_t$. Instead our \textsc{av} must estimate this state via observations from its sensors (see Fig (\ref{fig:scenario-with-tracking})). At time $t$, let us denote a full-snapshot of the world by $w_t$. This snapshot implicitly contains the relevant ground-truth state $x_t$, but also other scene information (e.g., vehicle types, dimensions etc.). A sensor $\mathcal{S}$ can be thought of as a function which takes $w_t$ and produces high-dimensional raw sensor data $\mathcal{S}(w_t)$ (e.g., a LiDAR point-cloud). This sensor data is then passed to a perception function $f$ (e.g., a neural-network obstacle detector \cite{qi2017pointnet++}), to produce an \emph{observation} $y_t \in \mathbb{R}^{d'}$ (e.g., the bounding boxes of other vehicles relative to the ego):

\begin{equation}
    y_t = f(\mathcal{S}(w_t))
    \label{eqn:observation}
\end{equation}
By using standard tracking algorithms \cite{thrun2002probabilistic}, we can use these observations to get an estimate $\hat{x}_t$ of the current state:

\begin{equation}
    \hat{x}_{t} = \mathds{E} \left[ x \mid y_{0 \dots t} \right]
    \label{eqn:track}
\end{equation}
If we have a one-step vehicle dynamics model $f_{\textit{dyn}}$, we can use it to predict the state in future time steps:

\begin{gather}
    \hat{x}_{t+i \mid t} = \underbrace{f_{\textit{dyn}} \circ \dots \circ f_{\textit{dyn}}}_{\text{$i$ times}}(\hat{x}_t)
    \label{eqn:prediction}
\end{gather}
Here, $\hat{x}_{t+i|t}$ denotes the predicted state at time $t$+$i$ given an estimate at $i$. In a real-world system, this setup allows us to sense, estimate, track and predict the state; in simulation, we have a problem: $f$ is typically a data-driven perception module trained on real sensors, but most simulators \emph{cannot generate} high fidelity sensor inputs (e.g., photo-realistic images). We can resolve this issue by re-framing our perception system as a noisy projection from the state space to observation space: 


\begin{equation}
    y_t = f(\mathcal{S}(w_t)) = H x_t + \epsilon(g(w_t)))
    \label{eqn:pem-approx}
\end{equation}
In this view, $f$ is composed of a $d' \times d$ projection matrix $H$ on state $x_t$, plus stochastic error dependent on the current world state $w_t$. The $\epsilon$ function is a surrogate model known as a \emph{Perception Error Model} \cite{piazzoni2023pem}. This is a probabilistic model of the original \textsc{av}'s perception noise, dependent on \emph{salient features} $g(w)$ extractable from simulated $w$. Salient features can include obstacle positions, dimensions, occlusion, or environment factors. We model $\epsilon$ as a \emph{gaussian process} \cite{williams2006gaussian},  where $m$, $\kappa$ are mean/kernel functions\footnote{The nuances of \textsc{gp}-inference and kernel choice are beyond the scope of this paper, but see \cite{duvenaud2014automatic} for discussion.}:

\begin{align}
   &\epsilon (w) \sim \textsc{GP}(m(g(w)), \kappa(g(w), g(w'))) 
   \label{eqn:gaussian-process}
\end{align}

Now, instead of using real-world sensor inputs directly, our simulator applies probabilistic noise from the \textsc{pem} based on the current simulated state. This makes each run of the simulator inherently stochastic, as different amounts of perceptual noise may be applied to observations on each run.


\subsection{Model Predictive Control for Highway Maneuvers}
\label{sec:method:mpc}

For a sufficiently complex control task such as lane changing, choosing the best actions $a_t$ at each time step is a \emph{non-linear constrained control} task. We phrase the controller of our \textsc{av} system under test as a \emph{Receding-Horizon Model-Predictive Control} optimization \cite{rawlings2017model}. Eq (\ref{eqn:npmc}) provides a formal definition of the optimization problem: At each time step $t$, the controller aims to choose actions $a_{t:(t+H)}$ over a finite time horizon $H$ which minimize a cost function $J(x, a)$. Actions must be chosen subject to a set of constraints $c_j(x,a)$ and obey physical dynamics $f_{dyn}$:

\newcommand{\sqd}[1]{\lVert #1 \rVert^2}

\begin{equation}
\begin{array}{@{}ll}
     \min\limits_{\substack{x_{t: (t + H)} \\ a_{t:(t+H)}}} & \sum\limits_{k=t}^{t+H} J(x_k, a_k) \\
     s.t. & x_{t} = \hat{x}_t  \\
     &\forall k,\ x_{k+1} = f_{\textit{dyn}}(x_{k}, a_{k}) \\
     &\forall j,\ c_j(x_k, a_k) < 0 \\
\end{array}
\label{eqn:npmc}
\end{equation}
Cost function $J(x,a)$ balances multiple factors such as tracking a reference velocity, minimizing abrupt movement, and staying close to the lane centre. The constraint functions $c_j(x,a)$ ensure states and actions remain feasible (e.g., that the car stays within road bounds and acceleration limits). 

We give a further breakdown of the cost function and implementation in Section (\ref{sec:exp:setup}). However, the purpose of describing Eq (\ref{eqn:npmc}) here in the context of our testing problem is to highlight that our \textsc{av}-controller represents yet another ``black-box'' component of our system: Despite behaving deterministically, solvers for nonlinear control problems are not guaranteed to find a global solution, and can perform arbitrarily poorly \cite{eiras2021two}. Other typical methods (such as reinforcement learning), pose the same problem.

\section{Estimating Specification Failure Probability}

We have defined our testing goal and outlined the components of our system-under-test. Now we can show how we formalize our safety properties, how we draw samples from the simulator, and how those aspects interact.

\label{sec:method}
\subsection{Specifying Safety with Signal Temporal Logic}
\label{sec:method:stl}

We can express \textsc{av} traffic rules involving statements about continuous values over time using Signal Temporal Logic \cite{maler2004monitoring}. \textsc{stl} has grammar:

\begin{equation}
    \begin{aligned}
        &\varphi := &\top \mid \eta \mid \neg \varphi \mid \varphi_1 \wedge \varphi_2 \mid \square_{I} \varphi \mid \\
        &&\Diamond_{I} \varphi \mid \varphi_1 \mathcal{U}_{I} \varphi_2  \mid H_{I} \varphi \mid O_{I} \varphi
    \end{aligned}
\label{eqn:stl-syntax}
\end{equation}

Here, $\eta$ is any predicate $\rho(x) - b \leq 0$ (with $b \in \mathbb{R}$ and function $\rho$ from state $x$ to $\mathbb{R}$). $\square_{I} \varphi$ means  $\varphi$ is \emph{always} true at every future time in interval $I$. $\Diamond_{I} \varphi$ means $\varphi$ must \emph{eventually} be true in $I$, $\varphi_{1} \mathcal{U}_{I} \varphi_{2}$ means $\varphi_1$ must remain true within $I$ \emph{until} $\varphi_{2}$ is true. $H_{I}$, $O_{I}$ are past versions of $\square_{I}$, $\Diamond_{I}$.

We can convert $\tau$ and $\varphi$ to a \emph{robustness metric} $\mathcal{L}(\varphi, \tau, i)$ over trajectories. This measures how strongly $\tau$ satisfied $\varphi$ (starting from time $i$). Large positive values indicate robust satisfaction; negative values a strong violation; near-zero values a trajectory on the boundary of satisfaction/violation. Eq (\ref{eqn:rob-semantics}) shows a subset of $\mathcal{L}$'s semantics. Other operators follow similar definitions \cite{deshmukh2017robust}.

\begin{equation}
\begin{array}{ll}
   \mathcal{L}(\rho(\tau) > 0, \tau, i) &=  \rho(x_i)\\
   \mathcal{L}(\neg \varphi, \tau, i) &= -\mathcal{L}(\varphi, \tau, i)\\
   \mathcal{L}(\varphi_1 \wedge \varphi_2, \tau, i) &= \min ( \mathcal{L}(\varphi_1, \tau, i), \mathcal{L}(\varphi_2, \tau, i) ) \\
   \mathcal{L}(\square_{I} \varphi, \tau, i) &= \inf\limits_{i' \in i + I} \mathcal{L}(\varphi, \tau, i')
\end{array}
   \label{eqn:rob-semantics}
\end{equation}

\subsection{Estimating the Rare Event with Splitting}
\label{sec:method:ams}

With our \textsc{pem}, \textsc{av}-controller, specification $\varphi$ and metric $\mathcal{L}$, we now describe our adaptive sampling contribution: Given $N$ simulated trajectories $\tau^{(1 \dots N)}$, we wish to calculate Eq (\ref{eqn:prob-failure-goal})---the probability a simulation violates $\varphi$. A naive approach might use \emph{Monte-Carlo} sampling:

\begin{equation}
    \mathds{E}\left[ \mathds{1}\{\mathcal{L}(\varphi, \tau, 0) < \gamma\} \right] \approx \frac{1}{N} \sum_{i=1}^{N} \mathds{1}\{\mathcal{L}(\varphi, \tau_i, 0) < \gamma\}
    \label{eqn:failure-probability}
\end{equation}
However, when violating $\varphi$ is rare, the number of simulations needed to achieve low relative error rapidly becomes infeasible \cite{juneja2006rare}. We instead take a \emph{multi-level splitting} approach \cite{cerou2019adaptive}. Rather than immediately estimating failure ($\gamma = 0$) we instead estimate decreasing thresholds $\gamma_1 > \gamma_2 > \dots > \gamma$. At each stage $m$, starting with $N$ trajectories we take two steps. First, we \emph{discard} trajectories that do not fall below threshold $\gamma_m$. Second, we \emph{replenish} back up to $N$ trajectories. To replenish discarded trajectories, we clone a random un-discarded $\tau^{(i)}$ up to $t'$---the first time step where $\mathcal{L}(\tau^{(i)}_{[0{:}t']}, \varphi, 0) < \gamma_m$. Then, we re-simulate $\tau^{(i)}$ from $t'$ to $T$. This ensures all $N$ trajectories at stage $m$ are below $\gamma_m$. With staged, partial re-samplings, Eq (\ref{eqn:prob-failure-goal}) becomes:

\begin{equation}
 \prod \limits_{m=1}^{M} P(\mathcal{L}(\varphi, \tau, 0) < \gamma_m | \mathcal{L}(\varphi, \tau, 0) < \gamma_{m-1}) 
\end{equation}
Given enough levels, each conditional probability should be significantly larger than $P(\mathcal{L}(\varphi, \tau, 0) < \gamma)$. The final computation: 

\begin{equation}
    \hat{p}_{ams} = \left\{ \prod^{M}_{m=1} \frac{N - K_m}{N} \right\} \times \frac{1}{N} \sum^{N}_{i=1} \mathds{1}\{\mathcal{L}(\varphi, \tau^{(i)}, 0)\}
    \label{eqn:ams-prob}
\end{equation}
has $M$ stages and $N$ initial simulations, where $K_m$ is the number discards per stage. Unlike adaptive importance sampling, \textsc{ams} guarantees convergence as $N \rightarrow \infty$ \cite{cerou2016Fluct}.

\subsection{Adaptive Splits via Online STL Monitors}
\label{sec:method:online-stl}

To achieve our high-level goal of adapting \textsc{ams} to \textsc{av} testing of \textsc{stl} specifications, we currently face a slight computational dilemma: Our robustness metric $\mathcal{L}(\tau, \varphi, 0)$ defines a single batch robustness value from the start to the end of a complete trajectory, and takes computation time proportional to the length of $\tau$. However in Section (\ref{sec:method:ams}), we saw that the discard and replenishment steps require access to the robustness values at \emph{arbitrary prefixes} of trajectories. In other words, \textsc{ams} re-simulation requires online computation of \emph{all partial trajectories}:

\begin{equation}
    \left\{ \mathcal{L}(\tau^{(i)}_{[0{:}t']}, \varphi, 0) \mid t' \in [0,T], i\in[0, N] \right\} 
    \label{eqn:partial-trajs}
\end{equation}
We resolve this dilemma by taking a key insight from the online monitoring literature---we can interleave partial computations into the \textsc{ams} process. To achieve this interleaving, we must first slightly alter our definition of $\mathcal{L}$ from (\ref{eqn:rob-semantics}) to define our metric in terms of partial rather than full trajectories. Eq (\ref{eqn:stl-nom}) is a modified metric $\mathcal{L}_{n}$ (where $n$ references the ``nominal semantics'' of \cite{deshmukh2017robust}, augmented with past operators). This definition makes explicit that we only partially evaluate trajectory $\tau$ up to fixed time step $t$: 
\newcommand{\Lnm}{\mathcal{L}_\textit{n}}

\begin{equation}
    \begin{array}{@{}l@{}l@{}}
         \Lnm (\rho(x)>0, \tau_{[0{:}t]}, i)\ \ &=\  \rho(\tau_{[i]})  \\
         \Lnm (\neg \varphi, \tau_{[0{:}t]}, i)\ &= - \Lnm (\varphi, \tau_{[0{:}t]}, i) \\
         \Lnm (\square_{I} \varphi, \tau_{[0{:}t]}, i)\ &= \inf\limits_{i' \in (i + I \cap [0, t])} (\Lnm (\varphi, \tau_{[0{:}t]}, i')) \\
         \Lnm (\Diamond_I \varphi, \tau_{[0{:}t]}, i)\ &= \sup\limits_{i' \in (i + I \cap [0, t])} (\Lnm (\varphi, \tau_{[0{:}t]}, i')) \\
         \Lnm (O_{I} \varphi, \tau_{[0{:}t]}, i)\ &= \sup\limits_{i' \in (i - I \cap [0, t])} (\Lnm (\varphi, \tau_{[0{:}t]}, i')) \\
         \Lnm (H_{I} \varphi, \tau_{[0{:}t]}, i)\ &= \inf\limits_{i' \in (i - I \cap [0, t])} (\Lnm (\varphi, \tau_{[0{:}t]}, i')) \\
         \Lnm (\varphi_1 \mathcal{U}_{I} \varphi_2, \tau_{[0{:}t]}, i)\ &= 
         \sup\limits_{i_2 \in (i + I \cap [0, t])} \min \left(
             \Lnm (\varphi_2, \tau_{[0{:}t]}, i_2), \vphantom{\inf \limits_{i_i \in [i, i_2]} } \right. \\
             \quad \left. \inf \limits_{i_1 \in [i, i_2]} \Lnm (\varphi_1, \tau_{[0{:}t]}, i_1) \right) \span
    \end{array}
\label{eqn:stl-nom}
\end{equation}
With the above definition, we could now naively compute all members of  (\ref{eqn:partial-trajs}) by re-evaluating $\varphi$ at every $i$, up to every $t$, at every re-sampling step. This is computationally wasteful, as the robustness values of many partial trajectories share many operations with the computations of their prefixes. To take advantage of this fact, we instead maintain a \emph{work-list} for each $\tau$. A \emph{work-list} stores a mapping from specification $\varphi$ and time step $t$ to robustness value $\mathcal{L}_n(\varphi, \tau_{[0{:}t]}, 0)$. By using a work-list, we can obtain a robustness value for the partial trajectories at time $t$+$1$ using just the newly available state $x_{t+1}$ and the previous values of the work-list, rather than repeating computations over the entire trajectory.

Alg (\ref{alg:wlist-update}) takes as input the current work-list at time step $t$ and sketches how it is updated online using the newly arrived state $x_{t+1}$:  For predicates $\rho(x)$, incoming state $x_{t+1}$ is added only if it is in $\varphi$'s \emph{time horizon}. For example, for $\varphi = \square_{[0,2]} (\rho(x) > 0)$, state $x_{3}$ would not be added, as it falls outside the relevant interval. For formulas like negation and conjunction, pointwise operators are leveraged to combine the existing results from previous sub-formula computations. Similarly for temporal operators, we can use the sliding min-max algorithm of \cite{lemire2006streaming} to compute running maxes over the relevant sub-formula intervals. Further optimizations can be added (e.g., replacing chunks of a work-list with `summaries' as sufficient information arrives \cite{deshmukh2017robust}) but we omit the details here. We can access the robustness of a partial trajectory from the updated work-list by querying $\textit{w-list}[\varphi][0]$. 

\begin{algorithm}
        \footnotesize
        \caption{Update Work List (Adapted from \cite{deshmukh2017robust})}
        \SetAlgoNoEnd
        \label{alg:wlist-update}
        \SetKwFunction{updwl}{upd-wl}
        \SetKwFunction{slmax}{sl-max}
        \SetKwFunction{rev}{reverse}
        \SetKwProg{myalg}{Function}{}{}
        \myalg{\updwl{w-list, $\varphi$, $x_{t+1}$}}{
            \Switch{$\varphi$}{
                \uCase{$\rho(x) > 0$}{
                    \lIf{$t$+$1$ is within time horizon of $\varphi$}{
                        $\textit{w-list}[\varphi][t+1] \gets \rho(x_{t+1})$ 
                    }
                }
                \uCase{$\neg \psi$}{
                    \updwl{\textit{w-list}, $\psi$, $x_{t+1}$} 
                    
                    $\textit{w-list}[\varphi] \gets$ Pointwise negation of $\textit{w-list}[\psi]$ 
                }
                \uCase{$\psi_1 \wedge \psi_2$}{
                    \updwl{\textit{w-list}, $\psi_1$, $x_{t+1}$}
                    
                    \updwl{\textit{w-list}, $\psi_2$, $x_{t+1}$} 
                    
                    $\textit{w-list}[\varphi] \gets$ Pointwise mins of $\textit{w-list}[\psi_1]$ and $\textit{w-list}[\psi_2]$ 
                }
                \uCase{$\square_{I} \psi$}{
                    \updwl{\textit{w-list}, $\psi$, $x_{t+1}$} 
                    
                    $\textit{w-list}[\varphi] \gets\ $Sliding min window of width $|I|$ across $\textit{w-list}[\psi]$
                }
                \uCase{$\Diamond_{I} \psi$}{
                    \updwl{\textit{w-list}, $\psi$, $x_{t+1}$} 
                    
                    $\textit{w-list}[\varphi] \gets\ $ Sliding max window of width $|I|$ across $\textit{w-list}[\psi]$
                }
                    
                \uCase{$\psi_1 \mathcal{U}_{I} \psi_2$}{
                    \updwl{\textit{w-list}, $\psi_1$, $x_{t+1}$}
                    
                    \updwl{\textit{w-list}, $\psi_2$, $x_{t+1}$} 

                    $\textit{lr-mins} \gets$ Pointwise mins of $\textit{w-list}[\psi_1]$ and $\textit{w-list}[\psi_2]$ 

                    \tcp{Calculate backwards inductively}
                    
                    \For{$i$ in descending timesteps}{
                        $\textit{us}[i] \gets $$\max \left(\textit{lr-min}[i], \min (\textit{w-list}[\psi_1][i], us[i+1])\right)$\DontPrintSemicolon
                    }
                    $\textit{w-list}[\varphi] \gets us$ 
                }
                \uCase{$H_{I} \psi$}{
                    \updwl{\textit{w-list}, $\psi$, $x_{t+1}$} 
                    
                    $\textit{w-list}[\varphi] \gets$ Sliding min window of width $|I|$ across (reversed) $\textit{w-list}[\psi]$
                }
                \uCase{$O_{I} \psi$}{
                    \updwl{\textit{w-list}, $\psi$, $x_{t+1}$} 
                    
                    $\textit{w-list}[\varphi] \gets$ Sliding max window of width $|I|$ across (reversed) $\textit{w-list}[\psi]$
                }
            }
        }
\end{algorithm}
\begin{algorithm}
        \footnotesize
        \SetAlgoNoEnd
        \caption{Online \textsc{stl-ams}}
        \label{alg:full-system}

        \SetKwFunction{stlams}{stl-ams}
        \SetKwFunction{updwl}{upd-wl}
        \SetKwProg{myalg}{Function}{}{}
        \myalg{\stlams{$x_0$, $\varphi$, $\gamma$, $T$, $K$, $N$}}{
            \For{$i \in [1,N], t \in [0,T]$}{
                $\epsilon_t, y_t, \hat{x}_t, a_t \gets$ Sample observations from PEM (\ref{eqn:pem-approx}-\ref{eqn:gaussian-process}), track via (\ref{eqn:track}) and choose actions by solving (\ref{eqn:npmc})
                
                Append $\langle x_t, a_t \rangle$ to trajectory $\tau^{(i)}$

                \tcp{Maintain work-list per trajectory (Alg \ref{alg:wlist-update})}
                
                $\textit{w-list}^{(i)}_{t+1} \gets$ \updwl{$\textit{w-list}^{(i)}_{t}$, $\varphi$, $x_{t+1}$}
                
                $L^{(i)}_{t+1} \gets \textit{w-list}^{(i)}_{t+1}[\varphi][0]$ \tcp{Robustness of $\tau^{(i)}_{[0:t+1]}$ }
                
                $x_{t+1} \gets$ Step forward simulation
            }
            
            Sort $\{L^{(0)}_T \dots L^{(N)}_T\}$ then set $\gamma_0 \gets L^{(K)}_T$

            $m \gets 0$
            
            \While{$\gamma_m > \gamma$}{
                $m \gets m + 1$
                
                Discard all trajectories trajectories $\tau^{(i)}$ where $L^{(i)}_{T} \geq \gamma_{k}$
                
                $\mathcal{I}_{k} \gets$ Indices of remaining un-discarded trajectories 
                
                \For{$i \in [0,N] \setminus \mathcal{I}_k$}{
                    Select a random $j \in {I}_k$

                    Find the first time step $t'$ where $L^{(j)}_{t'} < \gamma_k$
                    
                    $\tau^{(i)}_{[0{:}t']}, L^{(i)}_{[0{:}t']} \gets$ Copy values from $\tau^{(j)}$
                    
                    $\tau^{(i)}_{[t'{:}T]}, L^{(i)}_{[t'{:}T]} \gets$ Re-simulate $\tau^{(j)}$ from time $t'$
                }
                    
                Sort $\{L^{(0)}_T \dots L^{(N)}_T\}$ then set $\gamma_m \gets L^{(K)}_T$ 
            }
            \Return $\hat{p}_{ams}$ via Eqn (\ref{eqn:ams-prob})
        }
\end{algorithm}

Now that we can compute and cache \textsc{stl} robustness values for partial trajectories, we can interleave this with our \textsc{ams} sampler. Alg (\ref{alg:full-system}) provides an overview of our sampling technique for computing the probability that our \textsc{av} violates an \textsc{stl} specification in a stochastic simulation. It takes as input a starting state $x_0$, specification $\varphi$, failure threshold $\gamma$, initial simulation amount $N$ and discard rate $K$.

Lines (1-6) generate $N$ initial trajectories by simulating perceptual observations, control actions, and forward dynamics as outlined in sections (\ref{sec:method:pems}-\ref{sec:method:mpc}). Lines (7-8) track the \textsc{stl} robustness value of trajectories at every intermediate time step by maintaining up-to-date work-lists as described in Alg (\ref{alg:wlist-update}). Lines (9, 17) adaptively set discard thresholds $\gamma_m$ such that the $K$ safest trajectories are discarded at each stage. To replenish those discarded trajectories, lines (14-16) copy one of the remaining un-discarded $\tau^{(i)}$ up until the first time step $j$ where the robustness value of the partial trajectory falls below $\gamma_{m}$. We then re-simulate starting from $j$ to produce a new trajectory. We repeat this process until the discard threshold $\gamma_m$ falls below the desired failure threshold $\gamma$, then calculate a final estimate $\hat{p}_{\textit{ams}}$ using Eq (\ref{eqn:ams-prob}).

\section{Experiments}
\label{sec:exp:setup}

The following experiments use our running example of an \textsc{av} lane-change maneuver to evaluate our method. They demonstrate that Alg (\ref{alg:full-system}) can provide failure estimates for a full black-box \textsc{av}-system across multiple common traffic rules. When compared to baselines, we find Alg (\ref{alg:full-system}) provides more accurate failure estimates in fewer simulations. Further, we investigate how sampling performance differs across discard-rates and rule types. Fig (\ref{fig:scenario}) shows our CommonRoad simulation setup \cite{althoff2017commonroad}: The ego starts in the centre-lane at $15m$ with velocity $20\ m/s$. Its primary goal is to track a reference velocity of $v_g$=$30 m/s$. Three obstacles impede it---a static obstacle at $40m$ (forcing the ego to change lanes); A centre-lane vehicle at $50m$ with velocity $5\ m / s$, which cuts into the overtake lane after $0.6$ seconds (slowing the ego or forcing a lane change); a right-lane vehicle at $50m$, velocity $10m$, which merges after $1$ second (preventing the ego from slowing abruptly).\footnote{Full scenario specification and code at \url{github.com/craigiedon/CommonRules}} Simulations last $T$=$40$ steps ($4s$, $\Delta t$=$0.1s$). Dynamics evolve according to a kinematic single-track model \cite{rajamani2011vehicle}. 

Our \textsc{av}-system under test uses a standard pre-trained lidar-based obstacle detector---OpenPCDet's \emph{Multi-Head PointPillar} \cite{openpcdet2020}. As described in Section (\ref{sec:method:pems}), we train a surrogate \textsc{pem} to replicate the behaviour of this detector in simulation. The \textsc{pem} is composed of two separate Gaussian Processes: The first is a binary classifier, which predicts whether the lidar perception system would have failed to detect a given obstacle. The second is a regression model, which predicts how much noise would typically be added to the true location of an obstacle's bounding box. The \textsc{gp}s were fitted using \emph{Pyro} \cite{bingham2019pyro}, with an \textsc{rbf} kernel and sparse variational regression with 100 inducing points. As training data for fitting the \textsc{gp}s, we used $65500$ entries from the NuScenes Lidar Validation Set.\footnote{\url{https://www.nuscenes.org/nuscenes}} The \textsc{gp} input features were a 7-d vector---the x/y obstacle position, rotation, length/width/height dimensions, obstacle ``visibility category'' (where ``$1$'' means $\leq 40\%$ occlusion; ``$4$'' means $\geq 80\%$). The \textsc{gp} outputs consist of a 1-d binary variable for successful/unsuccessful obstacle detection, and a 3-d real-valued variable for the offsets between the obstacle's true x/y/rotation and the PointPillar estimate.

For tracking and predicting future vehicle locations, we used the interacting multiple models (\textsc{imm}) filter for lane-changes from \cite{carvalho2014stochastic}. This method operates similarly to a typical kalman filter, but instead of making estimates based on a single model, it maintains estimates from multiple models (i.e., for whether the vehicles will stay in the current lane, switch to the left lane, or switch to the right lane) and merges those estimates based on each model's current likelihood.

For model predictive control (Eq (\ref{eqn:npmc})) we use the lane-change controller from \cite{liu2017path}. At a high level, its cost function $J$ is comprised of 8 sub-goals: Reach a target destination; track a reference velocity, minimize acceleration, turn velocity, jerk and heading angle; stay close to the centre of the nearest lane; and avoid entering the ``potential field'' of other obstacles. To solve (\ref{eqn:npmc}), we used the Gurobi \cite{gurobi} optimizer. To ensure a feasible control action is always available, we first pre-solve a convex simplification of (\ref{eqn:npmc}) with CVXPy \cite{diamond2016cvxpy} (following \cite{eiras2021two}). For full implementation details of the cost-function sub-goals (and the weights used to balance them) see \cite{liu2017path} and the associated code for our paper.

\begin{table*}
    \caption{Interstate traffic rules (Predicate definitions in \cite{maierhofer2020formalization}).}
    \centering
    \ra{1.2}
    \scriptsize
    \begin{tabular}{@{}lL{6.0em}l@{}}
    Rule & Description & STL \\
    \toprule
    $\varphi_{1}$ & Safe Dist from Vehicles & 
        $\setlength{\jot}{1pt} 
        \begin{aligned}[t]
        &\square_{[0, \infty]}(\texttt{in\_same\_lane}(x^{\textit{ego}}, x^{o_i}) \wedge
        \texttt{in\_front\_of}(x^{\textit{ego}}, x^{o}) \wedge 
        \\
            &\ \neg O_{[0, t_{\textit{cut}}]}(
            \texttt{cut\_in}(x^{o}, x^{\textit{ego}}) \wedge H_{[1, \infty]}(\neg \texttt{cut\_in}(x^{o}, x^{\textit{ego}}))
        )  \\
        &\ \implies \texttt{keeps\_safe\_distance\_prec}(x^{\textit{ego}}, x^{o}))
        \end{aligned}$
    \\
    $\varphi_{2}$ & Unnecessary Braking & 
    
    $\square_{[0, \infty]} \left(\neg \texttt{unnecessary\_braking}(x^{\textit{ego}}, \{x^{o_1, \dots, o_3} \})\right)$ \\
    $\varphi_{3}$ & Preserve Traffic Flow &
    
    $\begin{aligned}[t]
    &\square_{[0, \infty]}(\neg \texttt{slow\_leading\_vehicle}(x^{\textit{ego}}, x^{o_1 \dots o_3}) \implies \texttt{preserves\_flow}(x^{\textit{ego}})) 
    \end{aligned}$
    \\
    
    $\varphi_{4}$ & Don't Drive Faster than Left Traffic &
    
    $\setlength{\jot}{1pt} 
    \begin{aligned}[t]
        &\square_{[0, \infty]}( \texttt{left\_of}(x^{o_i}, x^{\textit{ego}}) \wedge \texttt{drives\_faster}(x^{\textit{ego}}, x^{o_i}) \\
        &\implies (\texttt{in\_slow\_traffic}(x^{o_i}, x^{\{o_1, \dots o_3\} \setminus x^{o_i}}) \wedge \texttt{slightly\_higher\_speed}(x^{\textit{ego}}, x^{o_i})) \\
        &\quad \vee (\texttt{on\_access\_ramp}(x^{\textit{ego}}) \wedge \texttt{on\_main\_carriageway}(x^{o_i})) )
    \end{aligned}$ 
    \end{tabular} 
    \label{tab:stl-traffic-rules}
\end{table*}

We test our sampling method with respect to 4 formalizations of rules from the Vienna Convention on Road Traffic (taken from \cite{maierhofer2020formalization}). Table (\ref{tab:stl-traffic-rules}) shows the \textsc{stl} formulas for each rule. Full definitions of individual predicates (\texttt{in\_same\_lane}, \texttt{drives\_faster} etc.) are in \cite{maierhofer2020formalization}, but high level descriptions of each rule are as follows: $\boldsymbol{\varphi_1}$---maintain a minimum distance from vehicles in front (proportional to vehicle speed). If a vehicle ``cuts in'' from an adjacent lane, the ego gets $t_{\textit{cut}}$ seconds to re-establish distance. $\boldsymbol{\varphi_2}$---never drop acceleration below ``unnecessary'' levels (relative to vehicles in front). $\boldsymbol{\varphi_3}$---velocity should never fall below some minimum level (unless stuck in traffic). $\boldsymbol{\varphi_4}$---do not exceed the speed of left-lane vehicles unless merging from an access lane, or left-lane traffic is slow moving.

Our algorithm (listed as \textbf{STL-AMS} below) uses $N$=$250$ starting simulations, a discard amount of $K$=$25$, and final failure threshold of $\gamma=0$. We compare against three baselines: First, a Monte-Carlo sampler ($\textbf{Raw-MC}$), which runs $N$ simulations, estimating via (\ref{eqn:failure-probability}). Second, an importance sampler with a fixed proposal (\textbf{Imp-Naive}). We choose a proposal distribution which deliberately fails to detect 50\% of obstacles, and applies gaussian noise ($\mu$ = $0$, $\sigma^2$= $1$) to the bounding boxes of those it does detect. Third, a neural network-based importance sampler with an adaptive proposal learned via the \emph{cross-entropy method} (\textbf{Imp-CE}) from \cite{innes2023testing}. Its inputs and outputs are the same as those of the \textsc{gp}-\textsc{pem} described previously. Similar to \textsc{ams}, adaptive importance sampling proceeds in stages: At each stage $m$ (for a total of $M=10$ stages), $N_m$=$250$ trajectories are sampled and sorted by robustness.  \textbf{Imp-CE} then takes the lowest $K$=$0.1*N_m$ trajectories\footnote{Standard practice sets $K$ in range $[0.01, 0.2] * N_m$ \cite{corso2021survey}.} and minimizes their KL-divergence under the target \textsc{pem} versus current proposal. The intuition is that biasing our proposal towards the least robust trajectories in each stage should train a proposal which samples failure events with increasing probability.

\subsection{Results and Discussion}
\label{sec:exp:results}

\begin{table*}
    \centering
    \caption{Estimated Failure Probabilities (5 Repetitions)}
    \scriptsize
    \begin{tabular}{@{}lllll@{}}
        Method & $\varphi_1$ & $\varphi_2$ & $\varphi_3$ & $\varphi_4$   \\
        \toprule
        
        $\text{Raw-}\textsc{mc}_{250}$ & $1.2\text{e-}02\ (\pm 4.0\text{e-}03)$ & $0.0\ (\pm 0.0)$ & $0.0\ (\pm 0.0)$ & $0.0\ (\pm 0.0)$ \\
        
        
        Imp-Naive & $1.4\text{e-}08$ $(\pm 2.5\text{e-}08)$ & $2.8\text{e-}08$ $(\pm 8.4\text{e-}08)$ & $2.0\text{e-}08$ $(\pm 3.90\text{e-}08)$ & $0.0$ $(\pm 0.0)$ \\
        
        Imp-\textsc{ce} & $7.1\text{e-}06$ $(\pm 2.1\text{e-}05)$ & $7.6\text{e-}22$ $(\pm 2.2\text{e-}21)$ & $4.4\text{e-}18$ $(\pm 8.9\text{e-}18)$ & $0.0$ $(\pm 0.0)$ \\

        \textbf{\textsc{stl}-\textsc{ams} (ours)} & $\bm{8.8\textbf{e-}03}$ $\bm{(\pm 6.2\textbf{e-}03})$ & $\bm{1.5\textbf{e-}03}$ $\bm{(\pm 2.1\textbf{e-}03)}$ & $\bm{4.7\textbf{e-}03}$ $\bm{(\pm 3.4\textbf{e-}03)}$ & $\bm{3.5\textbf{e-}04}$ $\bm{(\pm 1.8\textbf{e-}03)}$ \\ 
        Ground Truth & $9.1\text{e-}03$ & $2.0\text{e-}03$ & $3.6\text{e-}03$ & $4.8\text{e-}05$
    \end{tabular}
    \label{tab:results-prob}
\end{table*}

Table (\ref{tab:results-prob}) shows estimated probabilities per rule. For ``ground-truth'', we computed Eq (\ref{eqn:failure-probability}) using $100000$ simulations. Across rules, our method produces the most accurate estimates.

\begin{figure}
    \centering
     \begin{subfigure}[t]{0.49\linewidth}
         \centering
         \includegraphics[width=\textwidth]{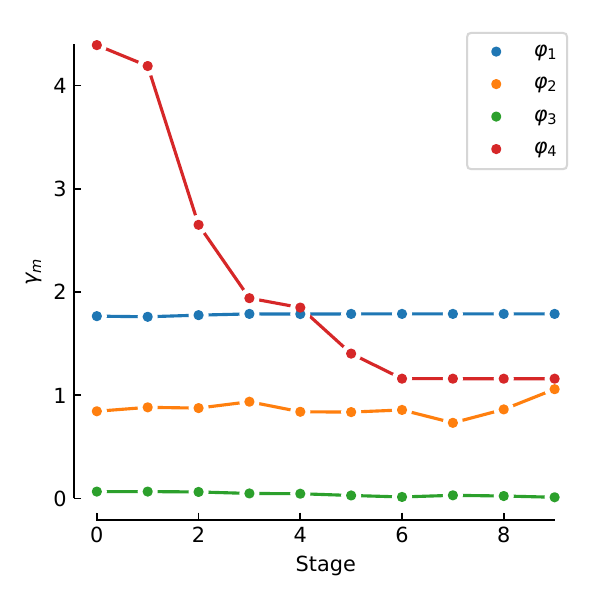}
         \caption{Robustness Threshold}
         \label{fig:ce-analysis:levels}
     \end{subfigure}
     \begin{subfigure}[t]{0.49\linewidth}
         \centering
         \includegraphics[width=\textwidth]{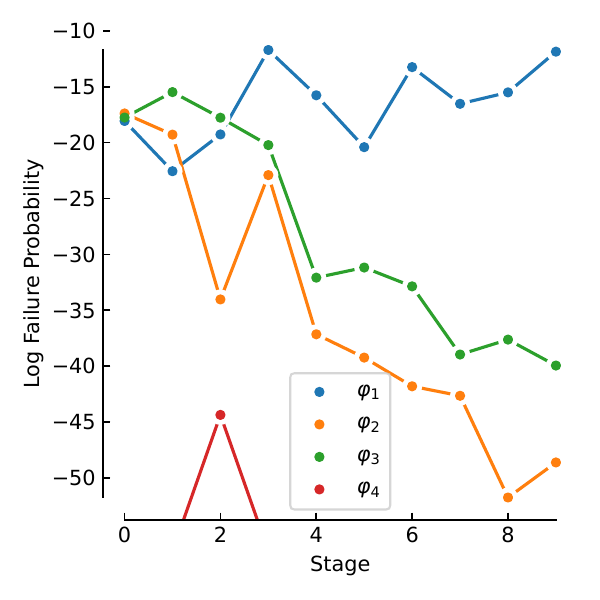}
         \caption{Log Probs (-$\infty$ below axis)}
         \label{fig:ce-analysis:log-prob}
     \end{subfigure}
     \begin{subfigure}[t]{0.49\linewidth}
         \centering
         \includegraphics[width=\textwidth]{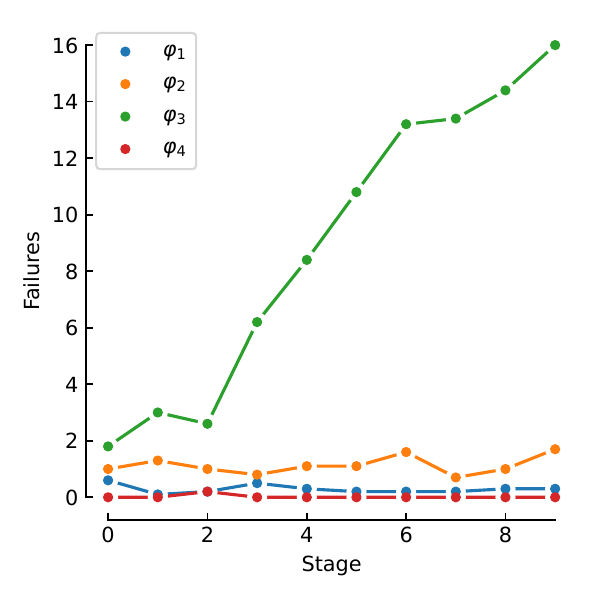}
         \caption{Failures}
         \label{fig:ce-analysis:failures}
     \end{subfigure}
    \caption{\textbf{Imp-CE} baseline performance over 10 stages of proposal learning.}
    \label{fig:ce-analysis}
\end{figure}

\textbf{Raw-MC} gives a reasonable estimate for $\varphi_1$ (the least rare). For other rules with lower probability, raw sampling yields zero simulation failures, resulting in $0\%$ estimates. \textbf{Imp-Naive} produces non-zero estimates for all specifications except $\varphi_4$, but vastly underestimates failure probability. This underscores the difficulty of fixed proposals---if a proposal distribution is too far from the original target, the likelihood weights per sample vary wildly, with estimates dominated by a tiny number of simulations. \textbf{Imp-CE} also produces unreliable estimates. Figs (\ref{fig:ce-analysis:levels}-\ref{fig:ce-analysis:failures}) provide insights why: As learning progresses for $\varphi_3$, the number of failures produced goes up, yet failure probability goes down. This suggests \textbf{Imp-CE} learns to bias its distribution towards a small set of unlikely failures, rather than approaching the true failure distribution. For $\varphi_4$, robustness thresholds initially decrease, but flatline around $m$=$6$---no failures found. We observe that this ``flatlining'' continues even past the $M=10$ stages documented in this paper. This highlights how challenging it can be to learn relevant features given longer horizons and state-spaces.

\begin{figure}
    \centering
     \begin{subfigure}[t]{0.48\linewidth}
         \centering
         \includegraphics[width=\textwidth]{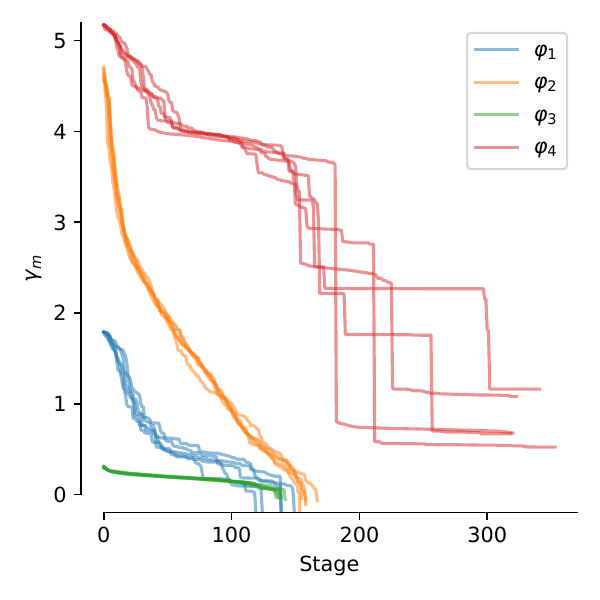}
         \caption{$K=2$}
     \end{subfigure}
     \begin{subfigure}[t]{0.48\linewidth}
         \centering
         \includegraphics[width=\textwidth]{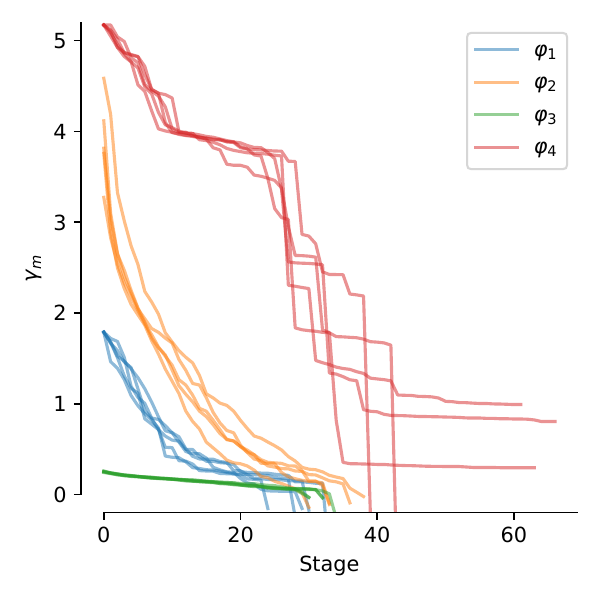}
         \caption{$K=25$}
     \end{subfigure}
     \begin{subfigure}[t]{0.48\linewidth}
         \centering
         \includegraphics[width=\textwidth]{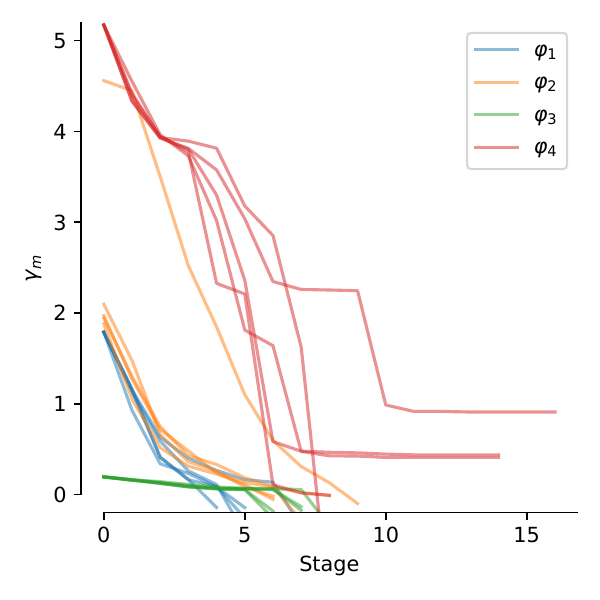}
         \caption{$K=125$}
     \end{subfigure}
     \begin{subfigure}[t]{0.48\linewidth}
         \centering
         \includegraphics[width=\textwidth]{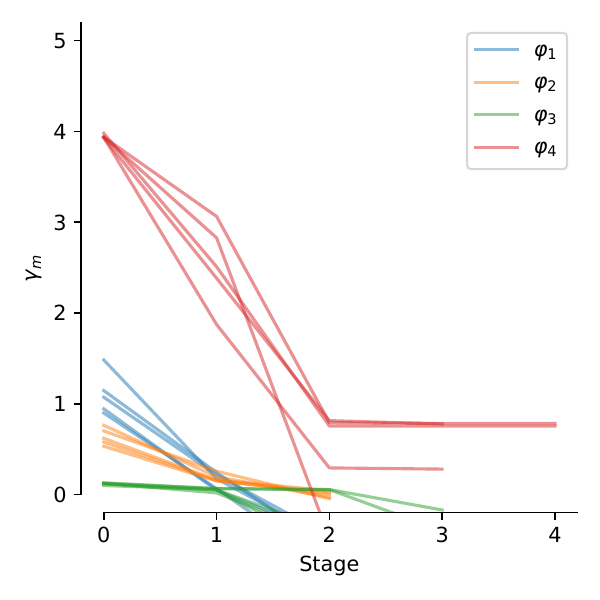}
         \caption{$K=225$}
     \end{subfigure}
    \caption{\textbf{STL-AMS} robustness thresholds by stage.}
    \label{fig:ams-levels}
\end{figure}

The results of Table (\ref{tab:results-prob}) are encouraging, but we found Alg (\ref{alg:full-system}) was sensitive to discard rate $K$. Fig (\ref{fig:ams-levels}) shows how threshold levels evolve at each stage (for $K$ values from $2$ to $225$). For $\varphi_4$ (the rarest), we found that too low or high $K$s caused unacceptable numbers of ``extinctions''\cite{cerou2019adaptive}---stages where all trajectories have identical robustness, rendering replenishment impossible.

Experiments demonstrate Alg (\ref{alg:full-system}) is viable for accurately estimating specification failure for a black-box \textsc{av}-system. However, this case study looks only at a single interstate traffic scenario, and our experiments necessarily have limitations: We considered perceptual disturbance as the sole source of simulation stochasticity; vehicle starting configurations remained fixed. Such experiments could be extended by placing a prior distribution over starts \cite{o2018scalable}, without altering the method. To trust our estimates, we also assume our simulator accurately represents reality. Whilst outside this paper's scope, clearly this assumption may not hold: Our \textsc{pem} may be an inaccurate surrogate of the test domain (i.e., it would be beneficial to incorporate work on \emph{\textsc{ml}-uncertainty calibration} \cite{guo2017calibration}). Our scenario also had fixed traffic behaviour, but real traffic is reactive and stochastic. Other works explore these issues in detail \cite{li2020interaction}. Finally, while it can be seen as an advantage that our method adaptively selects an appropriate number of simulations, and re-uses results from previous simulations, these advantages complicate comparisons of our method to others in terms of sample efficiency. In future work, we aim to compare performance across a wider range of scenarios in terms of \emph{fixed computational effort} across the full sampling pipeline.

\section{Related and Future Work}

This paper estimates failure probabilities. Similar tasks include falsification (find one failure) and adaptive stress testing (find the most-likely failure) \cite{corso2021survey}. Such tasks do not directly accomplish our goal, but may contain insights for rapidly guiding initial simulations towards failure areas. A related task is \emph{synthesis}---constructing controllers which explicitly obey $\varphi$ \cite{aasi2021control}. While synthesis can enforce adherence to specifications expressed in tractable \textsc{stl} subsets, the perceptual and control uncertainty in \textsc{av} scenarios means testing remains necessary.

Combining splitting and logic has been attempted previously \cite{jegourel2014effective}. Rather than use \textsc{stl} robustness, such works restrict themselves to heuristic decompositions of linear temporal logic formulae. This renders them unsuitable for cyber-physical domains like \textsc{av}.

One of our experiment baselines was importance sampling. Proposals are often represented by exponential distributions, since those have analytic solutions \cite{zhao2016accelerated}. Neural networks have also been used to represent the proposal (as we did), \cite{muller2019neural}. Most work considers rarity in the context of vehicle configurations or behaviour. Our work instead considers rarity in the context of perceptual disturbances. A sampler category unexplored in this paper are markov-chain methods \cite{botev2013markov}. With appropriate assumptions on metric smoothness and system linearizability, such techniques have shown promise in domains with long chains of dependent states \cite{sinha2020neural}. While out of scope, integrating such techniques with online \textsc{stl} monitoring may prove fruitful. 

Despite the asymptotic normality of \textsc{ams}, both splitting and sampling lack guarantees on estimation error for \emph{fixed} $N$. \emph{Certifiable sampling} addresses this with \emph{efficiency certificates} \cite{arief2021certifiable}---customized samplers with a bound on sampling error relative to $N$. However, certification methods depend on augmenting existing failure estimation algorithms, so techniques from this paper remain relevant.

Online and offline algorithms exist to calculate \textsc{stl} robustness \cite{donze2013efficient,deshmukh2017robust}. Typically, their efficiency is not considered in the context of a sampling regime. Yet recent advances in online monitoring could be leveraged within \textsc{ams} to discard infeasible trajectories early. For example, incorporating system dynamics, or causality \cite{yu2022online,deng2023causal}. 

Our experiments target interstate lane changes. Others encode rules for intersections, and situational awareness \cite{dokhanchi2018evaluating,maierhofer2022formalization}. In future work, we aim to assess sampling effectiveness across this diversity of specifications.




\bibliographystyle{splncs04}
\bibliography{references}

\end{document}